\documentclass{article}
\usepackage{ijcai21}

\usepackage{times}
\usepackage{soul}
\usepackage{url}
\usepackage[hidelinks]{hyperref}
\usepackage[utf8]{inputenc}
\usepackage[small]{caption}
\usepackage{graphicx}
\usepackage{amsmath}
\usepackage{amsthm}
\usepackage{booktabs}
\usepackage{algorithm}
\usepackage{algorithmic}
\usepackage{dsfont}
\usepackage{fixmath}
\usepackage{xcolor}
\usepackage{subcaption}
\usepackage{hyperref}   

\usepackage[left=2.1cm,right=2.1cm,bottom=2.2cm,top=2.2cm]{geometry}

\usepackage{xfp}
\usepackage{multirow}

\newtheorem{example}{Example}
\newtheorem{theorem}{Theorem}

\newtheorem{definition}{Definition}
\newtheorem{proposition}{Proposition}

\newcommand{\mat}[1]{\mathbold{#1}}

\title{Boolean Matrix Factorization with SAT and MaxSAT}

\author{
Florent Avellaneda
\and
Roger Villemaire
\affiliations
Université du Québec à Montréal, Department of Computer Science, Canada
\emails
\{avellaneda.florent, villemaire.roger\}@uqam.ca
}

\begin{document}

\maketitle

\begin{abstract}
The \emph{Boolean matrix factorization} problem consists in approximating a matrix by the Boolean product of two smaller Boolean matrices.
To obtain optimal solutions when the matrices to be factorized are small, we propose SAT and MaxSAT encoding; however, when the matrices to be factorized are large, we propose a heuristic based on the search for maximal biclique edge cover.
We experimentally demonstrate that our approaches allow a better factorization than existing approaches while keeping reasonable computation times. Our methods also allow the handling of incomplete matrices with missing entries.
\end{abstract}

\section{Introduction}

Boolean Matrix Factorization (BMF) is a fundamental problem in computer science.
This problem consists in representing a Boolean matrix as the Boolean product of two smaller matrices; this problem has applications in many fields such as in multi-label classification \cite{wicker2012multi}, role mining \cite{vaidya2007role} and bioinformatics \cite{liang2020bem}.
Due to its broad applications, BMF has recently attracted interest from the machine learning and AI communities \cite{DBLP:conf/ijcai/Miettinen020}.

Although many methods have been developed for the non-Boolean case, these methods fail to work in a Boolean context. For example, it is possible that the well-known \emph{Singular value decomposition} (SVD) obtains a greater reconstruction error than a BMF for the same decomposition size \cite{miettinen2008discrete}.
For this reason, specific algorithms dealing with the Boolean context must be developed.
The BMF problem is known to be NP-complete \cite{miettinen2008discrete}, and to the best of our knowledge, existing methods to solve it are heuristics.
Here, we propose using SAT and MaxSAT solvers to find optimal solutions to the BMF problem when datasets are small. Then, in the case of large datasets, we propose a heuristic approach based on the search for bicliques.

\section{Definitions}
We denote by $\mat{A}_{m \times n}$ a \emph{Boolean matrix} $\mat{A} \in \{0, 1\}^{n \times m}$ with $m$ rows, and $n$ columns and use $\mat{A}_{i, j}$ to represent the entry in the $i$-th row and the $j$-th column.
We also use the notation, $\mat{A}_{i, :}$ and $\mat{A}_{:, j}$, to represent the $i$-th row and the $j$-th column of $\mat{A}$, respectively.
A \emph{submatrix} $\mat{A}_{I, J}$ of $\mat{A}$ is the matrix obtained by selecting a subset $I \subseteq [1, m]$ of $\mat{A}$'s rows and a subset $J \subseteq [1, n]$ of $\mat{A}$'s columns.

Since the goal of this paper is to represent a Boolean matrix by the Boolean product of two smaller matrices, we start by defining this product.
\begin{definition}
    The \emph{Boolean product} of two matrices $\mat{A}_{m \times k}$ and $\mat{B}_{k \times n}$ is a matrix $(\mat{A} \circ \mat{B})_{m \times n}$ define by: 
    \[ (\mat{A} \circ \mat{B})_{i,j} = \bigvee_{\ell = 1}^k \mat{A}_{i,\ell} \mat{B}_{\ell,j}\]
\end{definition}
This definition is similar to the classical matrix product, where the product is replaced by the logical function ``AND" and addition by the logical function ``OR".

Note that $(\mat{A} \circ \mat{B})$ can also be written as the union of products of rank $1$ matrices 
    \[ (\mat{A} \circ \mat{B}) = \bigcup_{\ell = 1}^k( \mat{A}_{:,\ell} \circ \mat{B}_{\ell,:} ) \]
where $(\mat{A} \cup \mat{B})_{i,j} = \mat{A}_{i,j} \vee \mat{B}_{i,j}$.

The \emph{concatenation} of a matrix $\mat{A}_{m \times n}$ with a matrix $\mat{B}_{m \times n'}$ is the matrix $(\mat{A} | \mat{B})_{m \times (n+n')}$ such that $(\mat{A} | \mat{B})_{i,j} = \mat{A}_{i,j}$ for $j \leq n$ and $(\mat{A} | \mat{B})_{i,j}=\mat{B}_{i,j-n}$ for $j > n$.

We denote by $|\mat{A}|_1$ the number of $1$'s in $\mat{A}$, i.e, $\sum_{i,j} \mat{A}_{i,j}$. For a matrix $\mat{A}$, its \emph{transpose} is the matrix $\mat{A}^T$ with $(\mat{A}^T)_{i,j}=\mat{A}_{j,i}$.
For simplicity, the term ``matrix" will be used to describe a Boolean matrix.

\section{Optimal Boolean Matrix Factorization}

\subsection{Exact Factorization with SAT}\label{Exact Factorization with SAT}

In this section, we propose to factorize a matrix $\mat{X}_{m \times n}$ into two matrices $\mat{A}_{m \times k}$ and $\mat{B}_{k \times n}$ such that ${\mat{X} = \mat{A} \circ \mat{B}}$.
This factorization can be encoded in SAT as the set of following clauses.

For every $i, j$ such that $\mat{X}_{i,j} = 0$:
\begin{equation} \label{exact x=0}
    \bigwedge \limits_{\ell = 1}^k ( \neg \mat{A}_{i, \ell} \vee \neg \mat{B}_{\ell, j}  )
\end{equation}
These clauses guarantee that $(\mat{A} \circ \mat{B})_{i, j} = 0$.

For every $i, j$ such that $\mat{X}_{i,j} = 1$:

\begin{equation} \label{exact tmp var}
    \bigwedge \limits_{\ell = 1}^k ( \mat{T}_{i,j}^\ell \Rightarrow \mat{A}_{i,\ell} ) \wedge
    \bigwedge \limits_{\ell = 1}^k ( \mat{T}_{i,j}^\ell \Rightarrow \mat{B}_{\ell,j} )
\end{equation}
These clauses guarantee that if $\mat{T}_{i,j}^\ell = 1$ then $(\mat{A} \circ \mat{B})_{i, j} = 1$.

For every $i, j$ such that $\mat{X}_{i,j} = 1$:
\begin{equation} \label{exact x=1}
    \bigvee \limits_{\ell = 1}^k \mat{T}_{i,j}^\ell
\end{equation}
These clauses guarantee that one of the $\mat{T}_{i,j}^\ell$ is true for a given $i, j$ and therefore $(\mat{A} \circ \mat{B})_{i, j} = 1$.

Note that we can use this encoding to find an exact factorization into two minimum size matrices.
To do that, we can start with $k=1$ and increase $k$ while a solution is found.

\subsubsection{Symmetry Breaking}
Although the encoding we proposed in Section~\ref{Exact Factorization with SAT} is sufficient to realize the factorization, in this section we propose an optimization based on symmetry breaking.
Symmetry breaking, a well-known practice of the SAT community \cite{aloul2002solving,aloul2006efficient,brown1988backtrack}, consists in adding constraints to quickly remove some assignments as acceptable solutions.
The idea is to reduce the number of solutions that are equivalent to each other.

In our matrix factorization context, for any solution $\mat{A} \circ \mat{B}$ and for all $i, j \leq k$, we can switch the $i$-th column and the $j$-th column on the matrix $\mat{A}$ and switch the $i$-th row and the $j$-th row on the matrix $\mat{B}$ to obtain a new solution whose Boolean product remains unchanged.

To avoid this kind of permutation, we impose a lexicographical order on the rows of the matrix $\mat{B}$.
This can be performed with the following encoding. 

For every $i \in [1, m-1]$ and $j \in [1, n-1]$:
\begin{equation} \label{bs1}
    (\mat{Z}_{i,j} \wedge \mat{B}_{i,j} \wedge \mat{B}_{i+1, j}) \Rightarrow \mat{Z}_{i, j+1}
\end{equation}

For every $i \in [1, m-1]$ and $j \in [1, n-1]$:
\begin{equation} \label{bs2}
    (\mat{Z}_{i,j} \wedge \neg \mat{B}_{i,j} \wedge \neg \mat{B}_{i+1, j}) \Rightarrow \mat{Z}_{i, j+1}
\end{equation}

For every $i \in [1, m-1]$ and $j \in [1, n]$:
\begin{equation} \label{bs3}
    (\mat{Z}_{i, j} \wedge \neg \mat{B}_{i, j}) \Rightarrow \neg \mat{B}_{i+1, j}
\end{equation}

And finally, for each $i \in [1, m]$:
\begin{equation} \label{bs4}
    \mat{Z}_{i, 0}
\end{equation}

\subsubsection{Experimentation}

\begin{figure}
    \centering
    \includegraphics[width=0.9\linewidth]{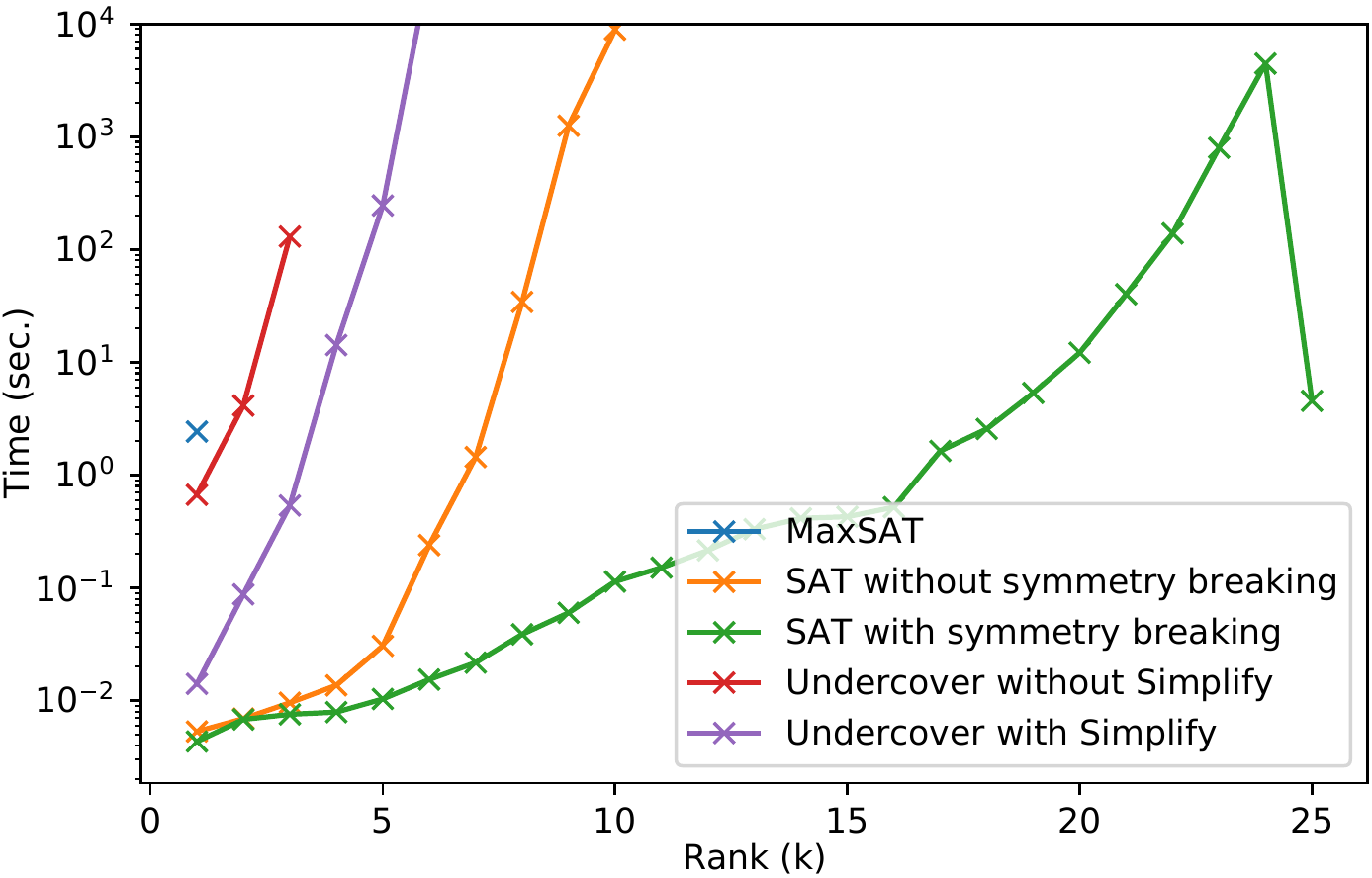}
    \caption{Execution time to factorize the Zoo dataset according to the desired rank}
    \label{fig:exact}
\end{figure}

We have implemented our encoding and used the SAT solver Kissat \cite{KISSAT}.
We performed several experiments on small real datasets, but in Figure~\ref{fig:exact} we only present the results on the Zoo dataset \cite{UCI}.

The first remark we can make is that adding the symmetry breaking constraints substantially improves computation times.
Second, we can see a notable difference between the computation time for $k=24$ and $k=25$.
In fact, it is much more difficult for the SAT solver to conclude that there is no exact factorization for $k=24$ than to find one for $k=25$.

\subsection{Factorization with MaxSAT} 
Even if there is no exact factorization of rank $k$, there can still be an approximate factorization yielding the right value for most of $\mat{X}$'s entries. Furthermore, in many applications, such as those where the data are noisy, an approximate factorization is the most we can expect from the data. As anticipated, most algorithms aim to produce an approximate factorization with as few errors as possible. In this section, we show how this problem can be tackled with the help of MaxSAT solvers.

MaxSAT solvers are an optimization extension of the SAT problem.
These solvers introduce soft variables: variables that should be assigned to true, but whose compliance with these assignments is not mandatory.
The aim of a MaxSAT solver is to find an assignment that satisfies a SAT formula and maximizes the number of satisfied soft variables.


To minimize the number of errors, i.e. the number of incorrect entries in the factorization $\mat{X}=\mat{A} \circ \mat{B}$, we could define $\mat{C}=(\mat{A}\circ \mat{B} \Leftrightarrow \mat{X})$, where $\mat{C}$ is a matrix of Boolean soft variables and $\Leftrightarrow$ is the if-and-only-if Boolean operator (equality) applied element-wise.
But, in fact, $\mat{C} \Rightarrow ( \mat{A} \circ \mat{B} \Leftrightarrow \mat{X})$ is sufficient since it assures that for each $\mat{C}_{i,j}=1$, $(\mat{A\circ B})_{i,j}=\mat{X}_{i,j}$, thereby, maximizing the number of $1$'s in $\mat{C}$, will minimize the number of errors in the factorization.
This brings the two following changes to the previous encoding.

First, for every $i, j$ such that $\mat{X}_{i,j} = 0$, the clauses (\ref{exact x=0}) are replaced by:

\begin{equation}
    \bigwedge \limits_{\ell = 1}^k ( \neg \mat{A}_{i, \ell} \vee \neg \mat{B}_{\ell, j} \vee \neg \mat{C}_{i, j} )
\end{equation}
These clauses reduce to (\ref{exact x=0})
when $\mat{C}_{i,j}=1$, therefore the number of $\mat{C}=1$ maximizes the number of entries for which (\ref{exact x=0}) holds.


Second, for every $i, j$ such that $\mat{X}_{i,j} = 1$, the clauses (\ref{exact x=1}) are replaced by:
\begin{equation}\label{relax exact x=1}
    \neg \mat{C}_{i, j} \vee \bigvee \limits_{\ell = 1}^k \mat{T}_{i,j}^\ell
\end{equation}
Again, these clauses reduce to (\ref{exact x=1}) when 
$\mat{C}=1$, and maximizing the number of $\mat{C}=1$ will maximize the number of entries for which (\ref{exact x=1}) holds.

\subsection{Experimentation} 

We have implemented our encoding and used the MaxSAT solver EvalMaxSAT \cite{avellaneda2020short}.
Our experiments show that results can be obtained in a reasonable amount of time only for tiny datasets and/or when the chosen rank is small.
For example, in Figure~\ref{fig:exact}, 2.3 seconds is required to find a factorization of rank 1 with a minimal number of errors, but many hours is not enough for a rank 2.
However, we observed that incomplete MaxSAT solvers performed well on small datasets with high quality factorizations. For instance, on the Zoo dataset, using the incomplete MaxSAT solver sls \cite{guerreiro2019sls}, we obtained better results than any existing factorization algorithm (accuracy of $\fpeval{round(100 * 274 / (101*28), 2)}$ \% for $k=3$, $\fpeval{round(100 * 123 / (101*28), 2)}$ \% for $k=7$ and $\fpeval{round(100 * 53 / (101*28), 2)}$ \% for $k=14$).
However, for larger datasets, using incomplete MaxSAT solvers yields lower quality results than existing approaches.

\section{Undercovering Boolean Matrix Factorization}

In this section, we focus on a particular case of approximate factorization that we call \emph{undercovering Boolean matrix factorization}.
This problem consists in finding a Boolean matrix factorization $\mat{D}=\mat{A} \circ \mat{B}$ containing as many of the $1$'s of $\mat{X}$ as possible. The key idea is that $\mat{D}$ does not introduce any false positives, i.e., a $1$ entry of $\mat{D}$ is a $1$ entry of $\mat{X}$.

\begin{definition}
    We say that the matrix $\mat{D}_{m \times n}$ \emph{undercovers} the matrix $\mat{X}_{m \times n}$ (denoted $ \mat{D} \leq \mat{X} $) if they are no $i, j$ such that $\mat{X}_{i,j} = 0$ and $\mat{D}_{i,j} = 1$.
\end{definition}


\begin{definition}
    We say that two matrices $\mat{A}_{m \times k}$ and $\mat{B}_{k \times n}$ form an optimal $k$-undercover of the matrix $\mat{X}_{m \times n}$ if:
    \begin{itemize}
        \item $(\mat{A} \circ \mat{B}) \leq \mat{X}$
        \item $\forall \mat{A}'_{m \times k}$ $\forall \mat{B}'_{k \times n}$: $(\mat{A}' \circ \mat{B}') \leq X \Rightarrow |\mat{A}' \circ \mat{B}'|_1 \leq |\mat{A} \circ \mat{B}|_1$ 
    \end{itemize}

\end{definition}

This problem has the advantage that it can be solved more efficiently with MaxSAT solvers than the general problem.
As we can see in Algorithm~\ref{alg:algorithm undercover}, each $0$ in the matrix to factorize, generates $2$-SAT clauses and no soft variables.
Moreover, we propose a method to further improve the resolution of this problem.
This method is represented in Algorithm 2 by the function $Simplify$ on line 11 and is explained in Subsection~\ref{generate card}.

\begin{algorithm}
\caption{Undercover}
\label{alg:algorithm undercover}
    \textbf{Input}: A matrix $\mat{X}_{m \times n}$ and an integer $k$.\\
    \textbf{Output}: A Boolean formula and a set of soft variables. 

    \begin{algorithmic}[1] 
    
    \STATE Initialize the formula $\Phi$ with constraints (\ref{bs1}), (\ref{bs2}), (\ref{bs3}), (\ref{bs4}).
    
    \FORALL{$(i, j) ~|~ \mat{X}_{i, j}=1$}
        \STATE $\Phi \leftarrow \Phi \wedge \bigwedge \limits_{\ell = 1}^k ( \mat{T}_{i,j}^\ell \Rightarrow \mat{A}_{i,\ell} )$ \hfill\COMMENT{Formula (\ref{exact tmp var})}
        \STATE $\Phi \leftarrow \Phi \wedge \bigwedge \limits_{\ell = 1}^k ( \mat{T}_{i,j}^\ell \Rightarrow \mat{B}_{\ell,j} )$ \hfill\COMMENT{Formula (\ref{exact tmp var})}
        \STATE $\mathcal{S} \leftarrow \mathcal{S} \cup \{ \neg \mat{C}_{i, j} \}$
        \STATE $\Phi \leftarrow \Phi \wedge ( \neg \mat{C}_{i, j} \vee \bigvee \limits_{\ell = 1}^k \mat{T}_{i,j}^\ell)$ \hfill\COMMENT{Formula (\ref{relax exact x=1})}
    \ENDFOR
    
    \FORALL{$(i, j) ~|~ \mat{X}_{i, j}=0$}
        \STATE $\Phi \leftarrow \Phi \wedge \bigwedge \limits_{\ell = 1}^k ( \neg \mat{A}_{i, \ell} \vee \neg \mat{B}_{\ell, j}  )$   \hfill\COMMENT{Formula (\ref{exact x=0})}
    \ENDFOR
    
    \RETURN $Simplify(\mat{X}, \Phi, \mathcal{S})$
    \end{algorithmic}
\end{algorithm}

\subsection{Generate Cardinalities} \label{generate card}

A recent and efficient algorithm for solving the MaxSAT problem is the OLL algorithm \cite{andres2012unsatisfiability}.
This algorithm originally created for ASP solvers has been adapted to MaxSAT solvers \cite{morgado2014core} and is used in many tools such as MSCG \cite{morgado2014mscg}, RC2 \cite{ignatiev2019rc2} and EvalMaxSAT \cite{avellaneda2020short}.

The principle of this algorithm is to search for unsatisfiable cores (sets of soft variables that cannot all be satisfied at the same time) in order to replace many soft variables by few cardinality constraints.
Here is an example to illustrate how this algorithm works.

\begin{example}
Let $\varphi_{soft} = \{x_1, x_2, x_3\}$ be a set of soft clauses and $\varphi_{hard} =\{(\neg x_1 \vee \neg x_2), (\neg x_2 \vee \neg x_3), (\neg x_1 \vee \neg x_3)\}$ be a set of hard clauses.

A call to a SAT solver on the set of clauses $ \varphi_{soft} \cup \varphi_{hard} $ will find that the formula is unsatisfiable and an unsatisfiable core can be $\{x_1, x_2\}$.
This mean that at least one of the two variables $x_1$ or $x_2$ has to be false and the cost of the MaxSAT formula is at least $1$.
Thus, we remove $x_1$ and $x_2$ from $\varphi_{soft}$,  we add the cardinally $(x_1 + x_2 \geq 1)$ to $\varphi_{soft}$ and we increment the cost of the formula.

On the second iteration, a SAT solver will find that $\varphi_{soft} \cup \varphi_{hard}$ is still unsatisfiable and the core will be $\{x_3, (x_1 + x_2 \geq 1)\}$. This mean that at least $x_3$ has to be false or $x_1 + x_2 \geq 2$.
Thus, we remove $x_3$ from $\varphi_{soft}$, replace $x_1 + x_2 \geq 1$ by $x_1 + x_2 \geq 0$, add a new cardinally $(x_3 + (x_1 + x_2 \geq 1) \geq 1)$ and increment the cost of the formula.

Now, the formula $ \varphi_{soft} \cup \varphi_{hard} $ is satisfiable and we can conclude that the cost of the initial formula is two.
\end{example}

We can notice that a call to a SAT solver is necessary to increment the cost of the formula.
Thus, if the cost of the formula to be solved is very high, it implies many calls to a SAT solver, which, in practice, can take a long time.

In this section we propose a method to significantly reduce the number of SAT calls to perform.
The idea is to use the knowledge of the problem modeled by the formula to be solved in order to quickly find sets of soft variables in which only $k$ of the variables can be satisfied at a time.
If at most $k$ variables can be satisfied in the set of variables $\{x_1, x_2,\ldots, x_f\}$, this means that we can increase the cost of the formula by $f-k$, removed the $f$ variables from the set $\varphi_{soft}$ and add a cardinality constraint $x_1 + x_2 + \cdots + x_n \geq k $ to $\varphi_{soft}$.

In order to find sets of soft variables for which only $k$ variables can be satisfied at the same time, we simply identify sets of $1$'s entries in the matrix to undercover that are two by two \emph{incompatible}.

\begin{definition}
    For a matrix $\mat{X}$, entries $\mat{X}_{i_1, j_1}=1$ and $\mat{X}_{i_2, j_2}=1$ are \emph{incompatible} if $\mat{X}$ contains either $\mat{X}_{i_1, j_2}=0$ or $\mat{X}_{i_2, j_1}=0$. 
\end{definition}

Now, the key property is that a rank-$1$ factorization can cover at most one of two incompatible $1$-entries.

\begin{proposition}\label{prop incompatible}
    Let $\mat{X}_{m \times n}$ be a matrix.
    If $\mat{X}_{i_1, j_1}=1$ is incompatible with $\mat{X}_{i_2, j_2}=1$ then for every product of rank-$1$ matrices $(\mat{A}_{m \times 1} \circ \mat{B}_{1 \times n}) \leq \mat{X}$, we have $(\mat{A} \circ \mat{B})_{i_1, j_1} = 0$ or $(\mat{A} \circ \mat{B})_{i_2, j_2} = 0$.
\end{proposition}

\begin{proof}
    If $(\mat{A} \circ \mat{B})_{i_1, j_1} = 1$ and $(\mat{A} \circ \mat{B})_{i_2, j_2} = 1$, then $\mat{A}_{i_1, 1} = 1$, $\mat{A}_{i_2, 1} = 1$, $ \mat{B}_{1, j_1} = 1$ and $ \mat{B}_{1, j_2} = 1$.
    This implies that $(\mat{A} \circ \mat{B})_{i_1, j_2} = 1$ and $(\mat{A} \circ \mat{B})_{i_2, j_1} = 1$ and therefore that $(i_1, j_1, 1)$ are not incompatible with $(i_2, j_2, 1)$.
\end{proof}

This furthermore implies a bound on how many entries in a set of two by two incompatible $1$ entries can be covered in a rank-$k$ factorization.

\begin{theorem}\label{thm amk}
    Let $\mat{X}_{m \times n}$ be a matrix.
    If $I \subseteq \mat{X}$ is a set of $1$'s entries in $\mat{X}$ two by two incompatible, 
    then at most $k$ of these entries can be covered by a $k$-undercover.
\end{theorem}

\begin{proof}
    By definition, a $k$-undercover correspond to two matrices $\mat{A}_{m \times k}$ and $\mat{B}_{k \times n}$ such that $\mat{A} \circ \mat{B} \leq \mat{X}$.
    Also by definition, $\mat{A} \circ \mat{B} = \bigcup_{\ell = 1}^{k} \mat{D}_\ell $ where $\mat{D}_\ell = \mat{A}_{:, \ell} \circ \mat{B}_{\ell, :}$.
    Since  $(\mat{A} \circ \mat{B}) \leq \mat{X}$ and $\mat{D}_\ell \leq (\mat{A} \circ \mat{B})$ we know that $\mat{D}_\ell \leq \mat{X}$ for each $\ell \in [1, k]$.
    By Proposition~\ref{prop incompatible}, since all elements from $I$ are two by two incompatible, each $\mat{D}_\ell$ can cover at most one element of $I$.
    Thus, at most $k$ elements of $I$ can be covered by a $k$-undercover.
\end{proof}

Algorithm~\ref{algo find atmostk} searches for sets of elements two by two incompatible in $\mat{X}$. 
For each set $I$ found, as we know from Theorem~\ref{thm amk} that at most $k$ elements can be covered by an undercover of rank $k$, the algorithm replaces the soft variables associated with elements from $I$ by a cardinality constraint.

\begin{algorithm}
\caption{Simplify\label{algo find atmostk}}
\label{alg:algorithm}
    \textbf{Input}: A matrix $\mat{X}_{m \times n}$, a Boolean formula $\Phi$ and a set of soft variables $\mathcal{S}$.\\ 
    \textbf{Output}: A simplified and equivalent Boolean formula and a set of soft variables. 
    \begin{algorithmic}[1] 
    \STATE $AllOne \leftarrow \{ (i, j)~|~\mat{X}_{i, j} = 1 \} $
    \WHILE{$true$}
        \STATE $toConsider \leftarrow AllOne$
        \STATE $noComp \leftarrow \{\}$
        \FORALL{$(i, j) \in AllOne$}
            \IF {$\forall (i', j') \in noComp: \mat{X}_{i, j'} = 0 \vee \mat{X}_{i', j} = 0$}
            \STATE $noComp \leftarrow noComp \cup \{(i, j)\} $
            \ENDIF
        \ENDFOR
        
        \IF {$noComp = \emptyset$}
            \RETURN $\Phi, \mathcal{S}$
        \ENDIF
        
        \STATE $AllOne \leftarrow AllOne \setminus noComp$
        
        \IF{$|noComp| > k$}
        
            \STATE $ \mathcal{S}' \leftarrow \{ \mat{D}_{i,j} ~|~ (i, j) \in noComp \}$
            \STATE $\Phi, card \leftarrow Cardinality( ( \sum \limits_{v \in \mathcal{S}'} v) \geq k)$
            \STATE $\mathcal{S} \leftarrow (\mathcal{S} \setminus \mathcal{S}') \cup \{card\}$
            
        \ENDIF
        
    \ENDWHILE
    \RETURN $\Phi, \mathcal{S}$
    \end{algorithmic}
\end{algorithm}

\begin{figure*}[h]
    \centering
    \begin{subfigure}[t]{0.33\textwidth}
        \includegraphics[width=\linewidth]{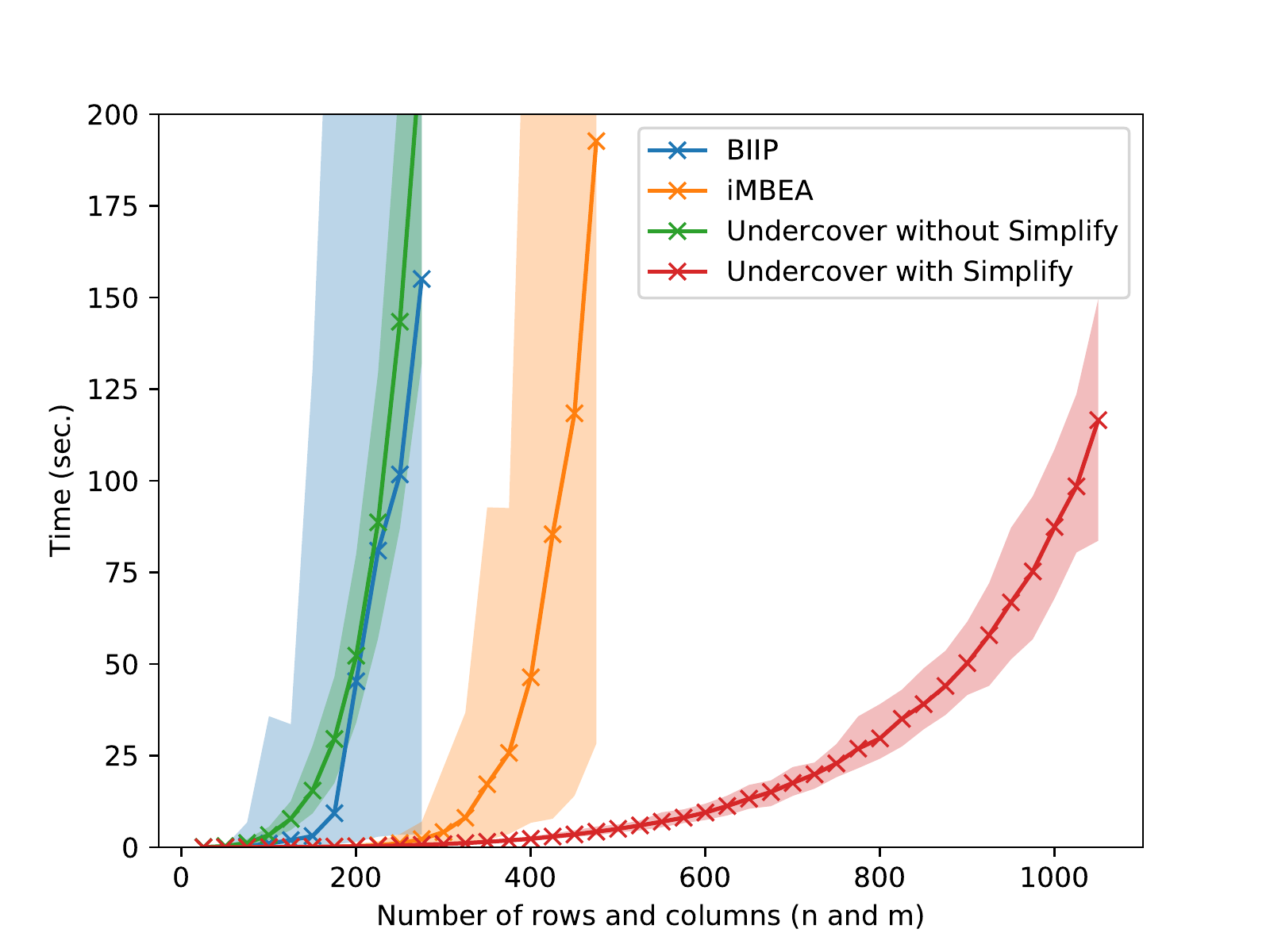}%
        \caption{ with $d=0.15$ and $k=10$.\label{bench eval size}}
    \end{subfigure}%
    \begin{subfigure}[t]{0.33\textwidth}
        \includegraphics[width=\linewidth]{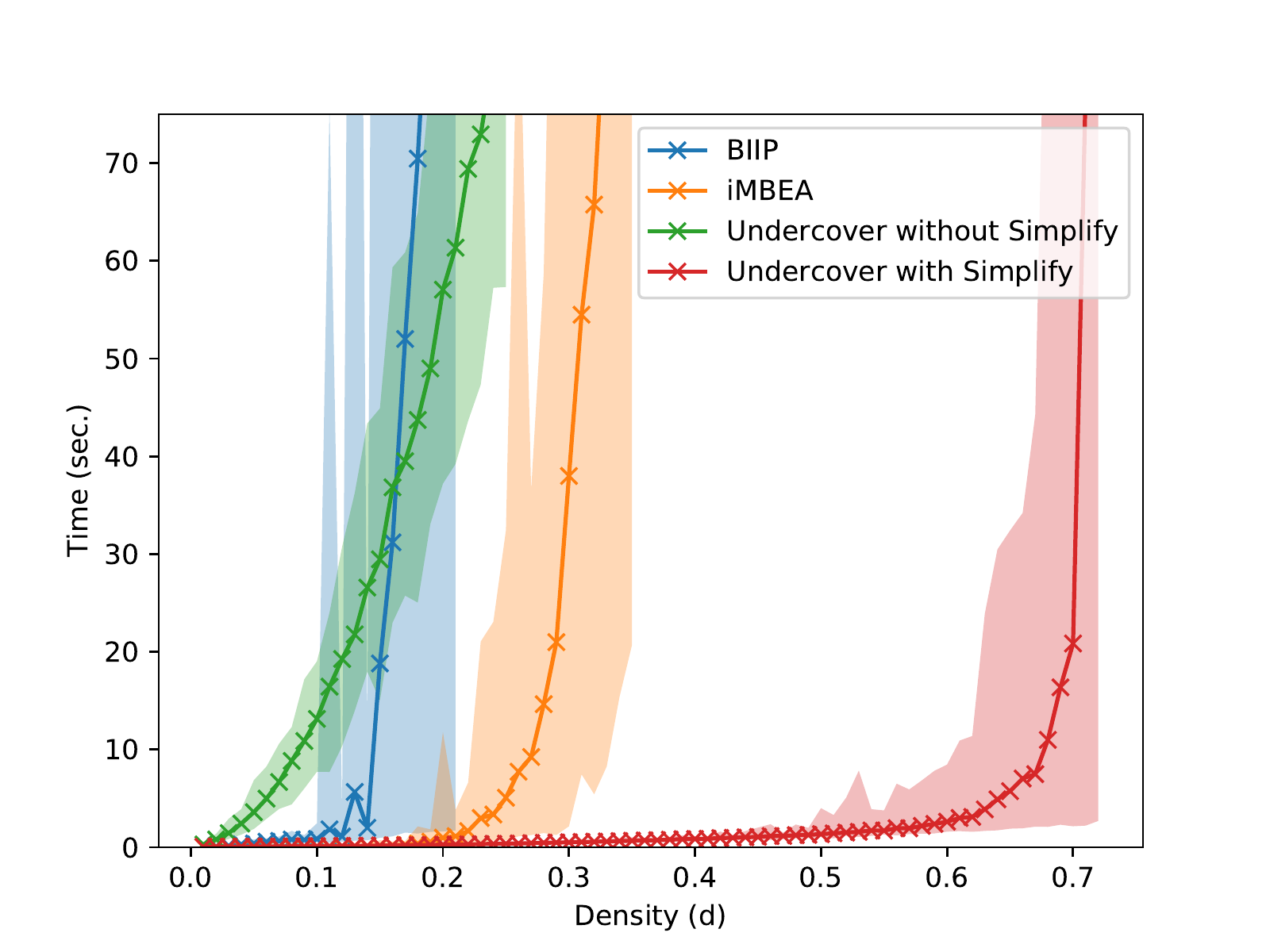}%
        \caption{with $k=10$, $n \times m = 175 \times 175$.\label{bench eval d}}
    \end{subfigure}%
    \begin{subfigure}[t]{0.33\textwidth}
        \includegraphics[width=\linewidth]{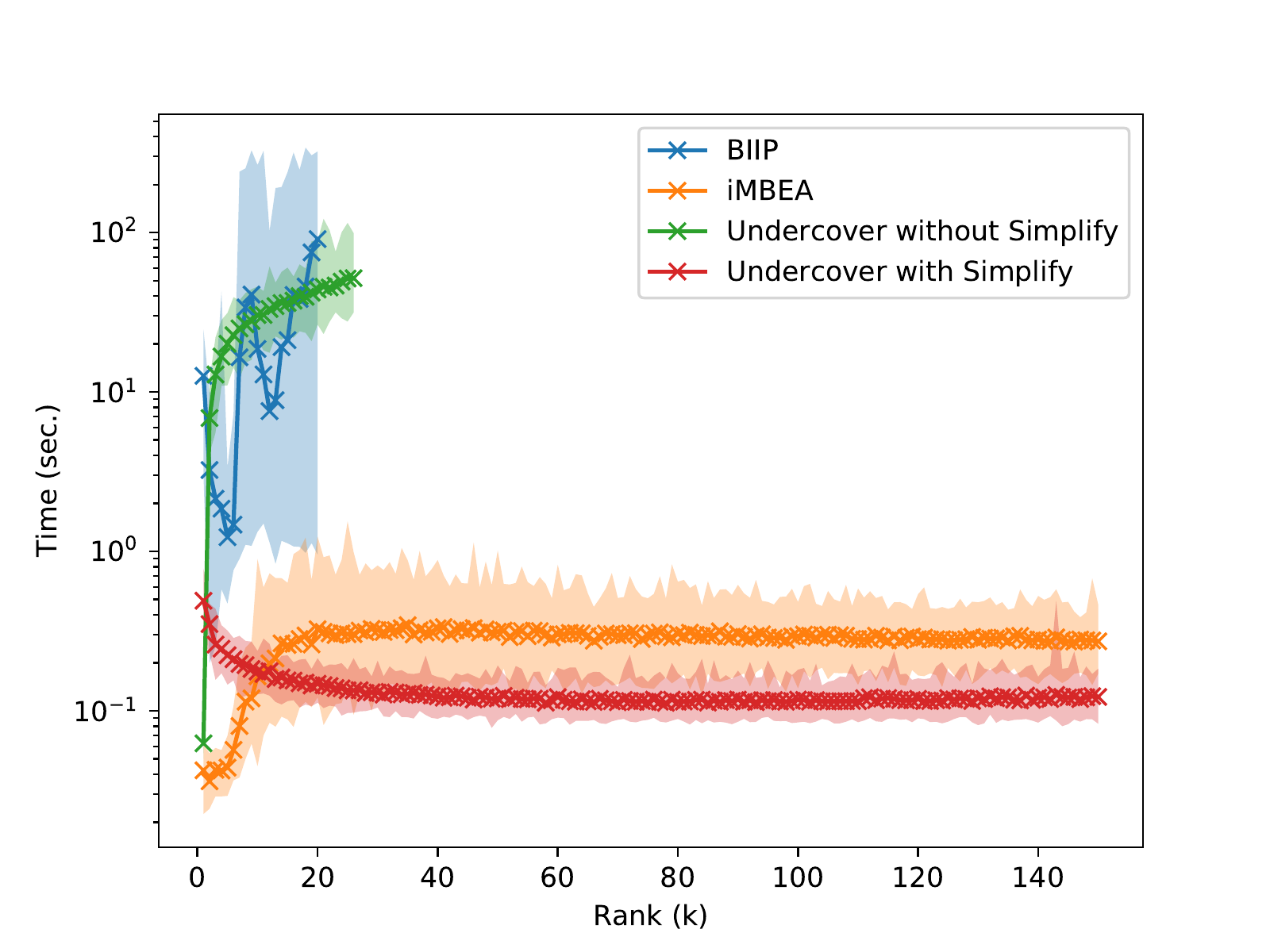}%
        \caption{with $d=0.15$, $n \times m = 175 \times 175$.\label{bench eval k}}
    \end{subfigure}%
    \caption{Average time, min and max time over 100 executions to find the largest biclique on a randomly generated matrix.}
\end{figure*}

\subsection{Related works}

A matrix $\mat{A}_{m \times n}$ can be seen as a bi-adjacency matrix of a bipartie graph $G = (U \cup V, E)$ where $U$ is the rows of $\mat{A}$, $V$ is the columns of $\mat{A}$ and $E = \{ \{i, j\} ~|~ \mat{A}_{i,j} = 1 \}$.
Thus, a $1$-undercover of $\mat{A}$ is equivalent to a edge biclique on $G$ and then an optimal $k$-undercover is a generalization of the problem of finding maximum edge biclique in bipartite graph.

This problem of finding a maximum edge biclique in a bipartite graph is a fundamental problem on the basis of several algorithms used in various fields of research.
For this reason, many methods have been proposed to solve this problem \cite{zhang2014finding,sozdinler2018finding,alexe2004consensus,prelic2006systematic}. 

To our knowledge, there are no tools to solve the k-undercover problem.
In the context of finding a maximum edge biclique in a bipartite graph (i.e., finding an optimal $1$-undercover on a matrix), we compare our method to two existing tools.
The first one is iMBEA \cite{zhang2014finding}, which, to our knowledge, is the most efficient tool to search for bicliques (we thank the authors for giving us access to their tool). The second tool is BIIP (publicly accessible online) which uses a similar approach to ours, i.e., the use of a constraints solver, in this case a commercial Integer Programming solver.

To compare the performance of these two tools to our approach with or without the cardinalities generation trick, we performed the following experiment.
We randomly generate two matrices $\mat{A}$ and $\mat{B}$ of rank $k$ 
and assigned the value $1$ to each cell with the probability of $\sqrt{ 1-(d^{1/k}) }$. Thus, ensuring that the probability of having a $1$ in a cell of $\mat{A} \circ \mat{B}$ is $d$. We then search for a $1$-undercover for this product.

We performed three benchmarks.
In Figure~\ref{bench eval size}, we fix the density $d$ to $0.15$ and rank $k$ to $10$ and evaluate the execution time according to the size of the matrix $\mat{A} \circ \mat{B}$.
In Figure~\ref{bench eval d}, we fix rank $k$ to $10$ and the size of the matrix $\mat{A} \circ \mat{B}$ to $175 \times 175$ and evaluate the execution time according to the density $d$.
Finally, in Figure~\ref{bench eval k}, we fix the density $d$ to $0.15$ and the size of the matrix $\mat{A} \circ \mat{B}$ to $175 \times 175$ and evaluate the execution time according to the rank $k$.
The choice of $d=0.15$, $k=10$ and $m=n=175$ was made because with these values, the four algorithms can be executed in a few seconds. Moreover, with these values, the execution time of \emph{UndercoverIT with Simplify} is equal to the execution time of \emph{iMBEA}. 

We notice that in our experiments, our algorithm \emph{Undercover with Simplify} is in most cases faster than other algorithms. 
The use of the ``simplify" method considerably improves the efficiency of our algorithm. 
However, it should be noted that the \emph{iMBEA} algorithm is faster than other algorithms when the density is very low or when $k$ is small. 

When comparing algorithms on real-world datasets, we found that our algorithm and the \emph{iMBEA} algorithm are complementary. Indeed, of the 11 datasets available
and used by the \emph{BIIP} algorithm in the benchmarks accompanying their paper, five are resolved faster by \emph{iMBEA}, four are resolved faster by our approach, and none are resolved faster by \emph{BIIP}.

\section{Efficient Boolean Matrix Factorization Heuristic}

To manage large datasets, we propose two heuristics in this section.
The idea of these heuristics is to solve the Boolean matrix factorization problem incrementally by solving a series of simpler problems.
In this case, we want to use the undercover problem to perform the factorization. 

The undercover problem is interesting because, in addition to having an effective way to solve it (as we saw in the previous section), an undercover guarantee that the cells with value $0$ on the matrix to be undercovered will not be covered by $1$.
Thus, thanks to this property, we can solve the factorization problem by making the union of undercover matrices.

\subsection{Relaxed Undercover Iteratively}

Our algorithm RUI (\emph{Relaxed Undercover Iteratively}), depicted on Algorithm~\ref{alg:Relaxed UndercoverIT}, consists in solving the problem by splitting it into sub-problems.
Thus, to approximate a $\mat{X}_{m \times n}$ matrix by the Boolean product of two matrices $\mat{A}_{m \times k}$ and $\mat{B}_{k \times n}$, the algorithm perform $it$ undercovers of rank $k'$.
After each undercover, for each covered $\mat{X}_{i,j}=1$, the soft variable $C_{i,j}$ is removed and covering this entry is hence no longer mandatory.
Note that since $\mat{X}_{i,j}=1$ is already covered, it is no longer necessary to cover this entry again since the concatenations of lines 4 and 5 will maintains the fact that $\mat{X}_{i,j}=1$ is covered by the product of $\mat{A}$ and $\mat{B}$.
%

Finally, the $relaxUndercover$ function on line 10 applies a simple greedy algorithm that consists in iteratively inverting the value of one of the cells of $\mat{A}$ or $\mat{B}$ to minimize $| (\mat{A} \circ \mat{B}) \oplus \mat{X} |_1$ as much as possible. Note that after the call to this function, $\mat{A} \circ \mat{B}$ may no longer be an undercover of $\mat{X}$.

\begin{algorithm}
\caption{Relaxed Undercover Iteratively}
\label{alg:Relaxed UndercoverIT}
    \textbf{Input}: A matrix $\mat{X}_{m \times n}$ and two integer $k'$ and $it$.\\
    \textbf{Output}: Two matrices $\mat{A}_{m \times (k' \cdot it)}$ and $\mat{B}_{(k' \cdot it) \times n} $ such that $\mat{A} \circ \mat{B} \approx \mat{X}$.
    \begin{algorithmic}[1] 
    
    \STATE $\Phi, \mathcal{S} \leftarrow undercover(\mat{X}, k' \cdot it)$
    
    \FOR{$i =  1$ \textbf{to} $it$}
    
        \STATE $\mat{A}', \mat{B}' \leftarrow MaxSAT(\Phi, \mathcal{S})$
        
        \STATE $\mat{A} \leftarrow (\mat{A} ~|~ \mat{A}')$
        
        \STATE $\mat{B} \leftarrow (\mat{B}^T~|~\mat{B}'^T)^T$
        
        \FORALL{$(i, j)$ such that $(\mat{A}' \circ
        \mat{B}')_{i,j} = 1$}
            \STATE $\mathcal{S} \leftarrow \mathcal{S} \setminus \{\mat{C}_{i,j}\}$
        \ENDFOR
    \ENDFOR
    \RETURN $\mat{A}, \mat{B} \leftarrow relaxUndercover(\mat{A}, \mat{B}, \mat{X})$
    \end{algorithmic}
\end{algorithm}

\subsection{Fast Relaxed Undercover Iteratively}

Since the execution time needed to perform an undercover increases significantly with the size of the matrix to be factorized, the idea of this second algorithm FRUI (\emph{Fast Relaxed Undercover Iteratively}) is to apply undercovering to sub-matrices.

Our approach involves finding an optimal $1$-undercover $\mat{A'}_{m \times 1}$ and $\mat{B'}_{1 \times n}$ with the constraint that a particular cell $\mat{X}_{i,j} = 1$ has to be covered by $\mat{A'} \circ \mat{B'}$.
If $\mat{X}_{i,j} = 1$ is covered by $\mat{A'} \circ \mat{B'}$, we know that for each $i'$ such that $\mat{X}_{i',j} = 0$, $\mat{A'}_{i', 1} = 0$ and for each $j'$ such that $\mat{X}_{i,j'} = 0$, $\mat{B'}_{1, j'} = 0$.
Therefore, it is not necessary to consider the entire $X$ matrix and it is sufficient to only keep rows in $\{i' ~|~ \mat{X}_{i',j} = 1\}$ and columns in $\{j' ~|~ \mat{X}_{i,j'} = 1\}$.
In practice, we choose $\mat{X}_{i,j}= 1$ as a cell to be covered such that ${|\mat{X}_{i, :}|_1 \times |\mat{X}_{:, j}|_1}$ is maximal.

\begin{algorithm}[t]
\caption{Fast Relaxed Undercover Iteratively}
\label{alg:Fast Relaxed UndercoverIT}
    \textbf{Input}: A matrix $\mat{X}_{m \times n}$ and an integer $k$.\\
    \textbf{Output}: Two matrices $\mat{A}_{m \times k}$, $\mat{B}_{k \times n}$ such that ${\mat{A} \circ \mat{B} \approx \mat{X}}$.
    
    \begin{algorithmic}[1] 
    
    \STATE $AlreadyCovered \leftarrow \emptyset$
    
    \FOR{$it = 1$ \textbf{to} $k$}
    
        \STATE Let $(i, j) \in \underset{(i, j)}{\arg\max}((\sum_{j'=1}^n \mat{X}_{i, j'}) \times (\sum_{i'=1}^m \mat{X}_{i', j}))$

        \STATE $keepRow \leftarrow \{i' ~|~ \mat{X}'_{i', j} = 1\}$
        
        \STATE $keepColumn \leftarrow \{j' ~|~ \mat{X}'_{i, j'} = 1\}$
        
        \STATE $\Phi, \mathcal{S} \leftarrow undercover(\mat{X}_{keepRow, keepColumn}, 1)$
        
        \STATE $\mathcal{S} \leftarrow \mathcal{S} \setminus AlreadyCovered$

        \STATE $\mat{A}' \leftarrow 0^{m \times 1} $
        \STATE $\mat{B}' \leftarrow 0^{1 \times n} $
        
        \STATE $\mat{A}'_{keepRow, 0}, \mat{B}'_{0, keepColumn} \leftarrow MaxSAT(\Phi, \mathcal{S})$

        \STATE $\mat{A} \leftarrow (\mat{A} ~|~ \mat{A}')$
        \STATE $\mat{B} \leftarrow (\mat{B}^T ~|~ \mat{B}'^T)^T$

        \FORALL{$(i, j)$ such that $(\mat{A}' \circ \mat{B}')_{i,j} = 1$}
            \STATE $AlreadyCovered \leftarrow AlreadyCovered \cup \{\mat{C}_{i,j}\}$
        \ENDFOR
    \ENDFOR
    
    \RETURN $relaxUndercover(\mat{A}, \mat{B}, \mat{X})$
    \end{algorithmic}
\end{algorithm}

\section{Benchmarks}

    \begin{table*}[h]
    \centering
    \begin{tabular}{llccccccccccc}
    \toprule
    \multirow{3}{*}{\textbf{Dataset}}  & \multirow{3}{*}{$\boldsymbol{k}$} & \multicolumn{2}{c}{\textbf{FRUI}} & \multicolumn{2}{c}{\textbf{RUI}} & \multicolumn{2}{c}{\textbf{Asso}} & \textbf{topFiberM} & \textbf{MEBF} & \textbf{GreConD} & \multicolumn{2}{c}{\textbf{bmad-XOR}} \\
    
    \cmidrule(lr){3-4}\cmidrule(lr){5-6}\cmidrule(lr){7-8}\cmidrule(lr){12-13}
    
    & & Error & Time  & Error & Time  & Error & Time & Error & Error & Error & Error & Time \\
    & & (\%) & (sec.) &  (\%) & (sec.) & (\%) & (sec.) & (\%) & (\%) & (\%) & (\%) & (sec.) \\
\midrule
    
    Lung
    
    & 3
    & \fpeval{round(100 * 997 / (31*147), 2)} & $\leq 0.1$
    & \fpeval{round(100 * 981 / (31*147), 2)} & 1.7
    & \fpeval{round(100 * 1164 / (31*147), 2)} & 9.7
    & \fpeval{round(100 * 1149 / (31*147), 2)} 
    & \fpeval{round(100 * 1260 / (31*147), 2)} 
    & \fpeval{round(100 * 1195 / (31*147), 2)} 
    & \textbf{\fpeval{round(100 * 966 / (31*147), 2)}} & 3.4 \\
    
    & 8 
    & \fpeval{round(100 * 773 / (31*147), 2)} & 0.1
    & \textbf{\fpeval{round(100 * 702 / (31*147), 2)}} & 12
    & \fpeval{round(100 * 1040 / (31*147), 2)} & 26
    & \fpeval{round(100 * 993 / (31*147), 2)} 
    & \fpeval{round(100 * 1123 / (31*147), 2)} 
    & \fpeval{round(100 * 950 / (31*147), 2)} 
    & \fpeval{round(100 * 793  / (31*147), 2)} & 4.1 \\
    
    & 16
    & \fpeval{round(100 * 497 / (31*147), 2)} & 0.1
    & \textbf{\fpeval{round(100 * 492 / (31*147), 2)}} & 12
    & \fpeval{round(100 * 687 / (31*147), 2)} & 57
    & \fpeval{round(100 * 566 / (31*147), 2)} 
    & \fpeval{round(100 * 906 / (31*147), 2)} 
    & \fpeval{round(100 * 634 / (31*147), 2)} 
    & \fpeval{round(100 * 607 / (31*147), 2)} & 4.6\\
    
    
    \midrule
    
    Phishing
    
    & 3
    & \fpeval{round(100 * 7794 / (1353*26), 2)} & 0.3
    & \textbf{\fpeval{round(100 * 7414 / (1353*26), 2)}} & 0.9
    & \fpeval{round(100 * 7598 / (1353*26), 2)} & 0.9
    & \fpeval{round(100 * 8227 / (1353*26), 2)} 
    & \fpeval{round(100 * 8281 / (1353*26), 2)} 
    & \fpeval{round(100 * 8620 / (1353*26), 2)} 
    & \fpeval{round(100 * 7527 / (1353*26), 2)} & 12\\
    
    & 6 
    & \fpeval{round(100 * 6276 / (1353*26), 2)} & 0.5
    & \textbf{\fpeval{round(100 * 6045 / (1353*26), 2)}} & 1.4
    & \fpeval{round(100 * 6569 / (1353*26), 2)} & 1.7
    & \fpeval{round(100 * 6249 / (1353*26), 2)} 
    & \fpeval{round(100 * 7666 / (1353*26), 2)} 
    & \fpeval{round(100 * 6757 / (1353*26), 2)} 
    & \fpeval{round(100 * 6216 / (1353*26), 2)} & 15\\
    
    & 13 
    & \fpeval{round(100 * 3135 / (1353*26), 2)} & 0.7
    & \textbf{\fpeval{round(100 * 3066 / (1353*26), 2)}} & 1.9
    & \fpeval{round(100 * 5085 / (1353*26), 2)} & 3.6
    & \fpeval{round(100 * 3681 / (1353*26), 2)} 
    & \fpeval{round(100 * 6794 / (1353*26), 2)} 
    & \fpeval{round(100 * 3411 / (1353*26), 2)} 
    & \fpeval{round(100 * 4114 / (1353*26), 2)} & 19\\

    \midrule

    Soybean
    
    & 10
    & \fpeval{round(100 * 3018 / (307*100), 2)} & 0.2
    & \textbf{\fpeval{round(100 * 2971 / (307*100), 2)}} & 3.6
    & \fpeval{round(100 * 3301 / (307*100), 2)} & 10
    & \fpeval{round(100 * 3317 / (307*100), 2)} 
    & \fpeval{round(100 * 4082 / (307*100), 2)} 
    & \fpeval{round(100 * 3332 / (307*100), 2)} 
    & \fpeval{round(100 * 3493 / (307*100), 2)}  & 17 \\
    
    & 25 
    & \fpeval{round(100 * 1791 / (307*100), 2)} & 0.3
    & \textbf{\fpeval{round(100 * 1660 / (307*100), 2)}} & 9.6
    & \fpeval{round(100 * 2560 / (307*100), 2)} & 26
    & \fpeval{round(100 * 2266 / (307*100), 2)} 
    & \fpeval{round(100 * 3389 / (307*100), 2)} 
    & \fpeval{round(100 * 1955 / (307*100), 2)} 
    & \fpeval{round(100 * 2739 / (307*100), 2)} & 27\\
    
    & 50 
    & \fpeval{round(100 * 655 / (307*100), 2)} & 0.4
    & \textbf{\fpeval{round(100 * 615 / (307*100), 2)}} & 12
    & \fpeval{round(100 * 2130 / (307*100), 2)} & 59
    & \fpeval{round(100 * 1308 / (307*100), 2)} 
    & \fpeval{round(100 * 2979 / (307*100), 2)} 
    & \fpeval{round(100 * 739 / (307*100), 2)} 
    & \fpeval{round(100 * 2010 / (307*100), 2)} & 42 \\
    
    
    \midrule

    Student
    & 18
    & \textbf{\fpeval{round(100 * 5864 / (395*176), 2)}} & 0.4
    & \fpeval{round(100 * 5898 / (395*176), 2)} & 31
    & \fpeval{round(100 * 6591 / (395*176), 2)} & 31
    & \fpeval{round(100 * 5891 / (395*176), 2)} 
    & \fpeval{round(100 * 8180 / (395*176), 2)} 
    & \fpeval{round(100 * 5910 / (395*176), 2)} 
    & \fpeval{round(100 * 6767 / (395*176), 2)} & 58 \\
    
    & 44 
    & \textbf{\fpeval{round(100 * 3387 / (395*176), 2)}} & 0.6
    & \fpeval{round(100 * 3397 / (395*176), 2)} & 69
    & \fpeval{round(100 * 5232 / (395*176), 2)} & 93
    & \fpeval{round(100 * 3521 / (395*176), 2)} 
    & \fpeval{round(100 * 7317 / (395*176), 2)} 
    & \fpeval{round(100 * 3400 / (395*176), 2)} 
    & \fpeval{round(100 * 5540 / (395*176), 2)} & 100\\
    
    & 88
    & \textbf{\fpeval{round(100 * 1198 / (395*176), 2)}} & 0.8
    & \fpeval{round(100 * 1234 / (395*176), 2)} & 71
    & \fpeval{round(100 * 4195 / (395*176), 2)} & 200
    & \fpeval{round(100 * 1610 / (395*176), 2)} 
    & \fpeval{round(100 * 6340 / (395*176), 2)} 
    & \fpeval{round(100 * 1227 / (395*176), 2)} 
    & \fpeval{round(100 * 3804 / (395*176), 2)} & 180\\
    
    \midrule

    Zoo
    
    & 3
    & \fpeval{round(100 * 355 / (101*28), 2)} & $\leq 0.1$
    & \textbf{\fpeval{round(100 * 274 / (101*28), 2)}} & $\leq 0.1$
    & \textbf{\fpeval{round(100 * 274 / (101*28), 2)}} & 0.9
    & \fpeval{round(100 * 337 / (101*28), 2)} 
    & \fpeval{round(100 * 294 / (101*28), 2)} 
    & \fpeval{round(100 * 345 / (101*28), 2)} 
    & \textbf{\fpeval{round(100 * 274 / (101*28), 2)}} & 2.9\\
    
    & 7 
    & \fpeval{round(100 * 157 / (101*28), 2)} & $\leq 0.1$
    & {\fpeval{round(100 * 150 / (101*28), 2)}} & $\leq 0.1$
    & \fpeval{round(100 * 168 / (101*28), 2)} & 2.0
    & \fpeval{round(100 * 191 / (101*28), 2)} 
    & \fpeval{round(100 * 214 / (101*28), 2)} 
    & \fpeval{round(100 * 178 / (101*28), 2)} 
    & \textbf{\fpeval{round(100 * 132 / (101*28), 2)}} & 3.7\\
    
    & 14
    & \textbf{\fpeval{round(100 * 63 / (101*28), 2)}} & $\leq 0.1$
    & \textbf{\fpeval{round(100 * 63 / (101*28), 2)}} & $\leq 0.1$
    & \fpeval{round(100 * 143 / (101*28), 2)} & 3.8
    & \fpeval{round(100 * 109 / (101*28), 2)} 
    & \fpeval{round(100 * 161 / (101*28), 2)} 
    & \textbf{\fpeval{round(100 * 63 / (101*28), 2)}} 
    & \fpeval{round(100 * 88 / (101*28), 2)} & 3.2\\
    

\bottomrule
\end{tabular}
    \caption{Benchmark results.\label{tab: main result}}
    \end{table*}

To compare our approach with existing tools, we performed benchmarks on five real-world datasets, but benchmarks on 18 datasets can be found in the annex.
We used classical state of the art datasets from UCI \cite{UCI}.
The datasets are binarized with a one-shot encoding and characteristics of each dataset are described on Table~\ref{tab: datasets}. 
A factorization for each dataset is performed with three different rank values: $10 \%$, $25 \%$ and $50 \%$ of $min(rows, columns)$.
We compared RUI (Algorithm~\ref{alg:Relaxed UndercoverIT}) and FRUI (Algorithm~\ref{alg:Fast Relaxed UndercoverIT}) with $k'=1$ and $it=k$ versus Asso \cite{miettinen2008discrete}, topFiberM \cite{desouki2019topfiberm}, MEBF \cite{wan2020fast}, GreConD \cite{belohlavek2010discovery}, bmad-XOR \cite{wicker2019xor} and present the results in Table~\ref{tab: main result}.
Note that bmad-XOR solves a different problem since it uses an ``XOR" operator instead of an ``OR" operator for the Boolean product. Since bmad-XOR uses randomness in their algorithm, the quality of the results greatly varies from one run to the next, so we ran their algorithm 1000 times and report the best run and the time used to run all executions. 
Our algorithms were implemented in C++ using the MaxSAT solver EvalMaxSAT.
All algorithms were executed on Ubuntu 20.04 with Intel\textsuperscript{\textregistered} Core\texttrademark ~CPU i7-2600K @ 3.40GHz and 8 GB of RAM.

We see that the factorizations found by our methods are almost always of better quality than the solutions found by the classical methods.
We also see that the ``fast" version of our approach greatly improves executing times with minimal impact on the quality of the solutions found.

\begin{table}
\centering
\begin{tabular}{lccc}
\toprule
\textbf{Dataset} & \textbf{Rows} & \textbf{Columns} & \textbf{Density}\\
\midrule
Lung        & 31        & 147   & \fpeval{round( (1486) / (1486+3203), 2)} \\
Phishing    & 1353      & 26    & \fpeval{round( (11507) / (+1353*26), 2)} \\
Soybean     & 307       & 100   & \fpeval{round( (6315) / (6315+22897), 2)}\\
Student     & 395       & 176   & \fpeval{round( (9254) / (9254+60266), 2)}\\
Zoo         &  101      & 28    & \fpeval{round( (640) / (640+2188), 2)} \\
\bottomrule
\end{tabular}
\caption{Numbers of rows, number of columns and fraction of $1$'s.\label{tab: datasets}}
\end{table}

\section{Conclusion}
We proposed SAT and MaxSAT encodings to find optimal solutions to two problems: the exact factorization of a matrix into two minimum size matrices and the optimal approximation of a matrix by the Boolean product of two fixed size matrices.
We experimentally showed that it is possible to find optimal factorization for small datasets.
Note that since our methods consists in generating constraints for each entries present in the matrices to factorize, it is straightforward to adapt our approaches to matrices with missing entries.

For the case where datasets are too large to find an optimal solution, we proposed a method based on an undercover search.
Although the undercover problem is a generalization of the maximum biclique search problem in a bipartite graph, we showed experimentally that our method can find maximum biclique faster in most cases than the best algorithms of the specialized literature on the biclique search problem.

Finally, we showed that our approximation heuristics based on this undercover search generally produced factorization of better quality than classical literature techniques while maintaining reasonable computation times.

\newpage
\bibliographystyle{named}
\bibliography{ijcai21}

\newgeometry{left=0.5cm,top=0.5cm,bottom=0.1cm,right=0.1cm}

    \begin{table}[ht]
    \centering
    \begin{tabular}{|l||l||c|c||c|c||c|c||c|c||c|c||c|c||c|c|}
    \hline
    \multirow{3}{*}{dataset}  & \multirow{3}{*}{$k$} & \multicolumn{2}{c||}{FRUI} & \multicolumn{2}{c||}{RUI} & \multicolumn{2}{c||}{Asso} & \multicolumn{2}{c||}{topFiberM}& \multicolumn{2}{c||}{MEBF} & \multicolumn{2}{c||}{GreConD} & \multicolumn{2}{c|}{bmad-xor} \\
    & & Error & Time  & Error & Time  & Error & Time & Error & Time & Error & Time & Error & Time & Error & Time \\
    & & (\%) & (sec.) &  (\%) & (sec.) & (\%) & (sec.) & (\%) & (sec.)& (\%) & (sec.) & (\%) & (sec.) & (\%) & (sec.) \\
    \hline
    \hline
    
    \multirow{3}{*}{Audio}
    
    & 20
    & \fpeval{round(100 * 820 / (200*310), 2)} & 0.3
    & \textbf{\fpeval{round(100 * 798 / (200*310), 2)}} & 1.2
    & \fpeval{round(100 * 1146 / (200*310), 2)} & 76
    & \fpeval{round(100 * 972 / (200*310), 2)} & 0.1
    & \fpeval{round(100 * 1621 / (200*310), 2)} & 0.3
    & \fpeval{round(100 * 871 / (200*310), 2)} & 0.1
    & \fpeval{round(100 * 1214 / (200*310), 2)} & 37\\
    
    & 50
    & \textbf{\fpeval{round(100 * 375 / (200*310), 2)}} & 0.5
    & \fpeval{round(100 * 383 / (200*310), 2)} & 2.0
    & \fpeval{round(100 * 881 / (200*310), 2)} & 170
    & \fpeval{round(100 * 702 / (200*310), 2)} & 0.1
    & \fpeval{round(100 * 1418 / (200*310), 2)} & 0.2
    & \fpeval{round(100 * 393 / (200*310), 2)} & 0.2
    & \fpeval{round(100 * 976 / (200*310), 2)} & 64\\
    
    & 100
    & \textbf{\fpeval{round(100 * 195 / (200*310), 2)}} & 1.0
    & \fpeval{round(100 * 200 / (200*310), 2)} & 3.3
    & \fpeval{round(100 * 642 / (200*310), 2)} & 320
    & \fpeval{round(100 * 594 / (200*310), 2)} & 0.3
    & \fpeval{round(100 * 1418 / (200*310), 2)} & 0.2
    & \fpeval{round(100 * 198 / (200*310), 2)} & 0.3
    & \fpeval{round(100 * 765 / (200*310), 2)} & 104\\

    \hline

    \multirow{3}{*}{Breast}
    
    & 9
    & \fpeval{round(100 * 3136 / (699 * 90), 2)} & 0.2
    & \fpeval{round(100 * 2945 / (699 * 90), 2)} & 1.9
    & \fpeval{round(100 * 3056 / (699 * 90), 2)} & 8.0
    & \textbf{\fpeval{round(100 * 2928 / (699 * 90), 2)}} & $\leq 0.1$
    & \fpeval{round(100 * 4152 / (699 * 90), 2)} & 0.1
    & \fpeval{round(100 * 3127 / (699 * 90), 2)} & $\leq 0.1$
    & \fpeval{round(100 * 3201 / (699 * 90), 2)} & 23 \\
    
    & 22 
    & \textbf{\fpeval{round(100 * 1814 / (699 * 90), 2)}} & 0.3
    & \fpeval{round(100 * 1986 / (699 * 90), 2)} & 2.8
    & \fpeval{round(100 * 2401 / (699 * 90), 2)} & 20
    & \fpeval{round(100 * 2113 / (699 * 90), 2)} & 0.1
    & \fpeval{round(100 * 3993 / (699 * 90), 2)} & 0.2
    & \fpeval{round(100 * 2032 / (699 * 90), 2)} & $\leq 0.1$
    & \fpeval{round(100 * 2672 / (699 * 90), 2)} & 38\\
    
    & 45 
    & \textbf{\fpeval{round(100 * 853 / (699 * 90), 2)}} & 0.3
    & \fpeval{round(100 * 1017 / (699 * 90), 2)} & 3.5
    & \fpeval{round(100 * 1771 / (699 * 90), 2)} & 40
    & \fpeval{round(100 * 1242 / (699 * 90), 2)} & 0.1
    & \fpeval{round(100 * 3745 / (699 * 90), 2)} & 0.4
    & \fpeval{round(100 * 1021 / (699 * 90), 2)} & $\leq 0.1$
    & \fpeval{round(100 * 1776 / (699 * 90), 2)}  & 59
    \\

    \hline

    \multirow{3}{*}{Chess}
    
    & 4
    & \fpeval{round(100 * 18479 / (3196*39), 2)} & 1.0
    & \textbf{\fpeval{round(100 * 16681 / (3196*39), 2)}} & 3.7
    & \fpeval{round(100 * 17054 / (3196*39), 2)} & 2.0
    & \fpeval{round(100 * 21540 / (3196*39), 2)} & $\leq 0.1$
    & \fpeval{round(100 * 20560 / (3196*39), 2)} & 0.3
    & \fpeval{round(100 * 18064 / (3196*39), 2)} & $\leq 0.1$
    & \fpeval{round(100 * 16978 / (3196*39), 2)} & 37 \\
    
    & 10 
    & \fpeval{round(100 * 10681 / (3196*39), 2)} & 1.6
    & \textbf{\fpeval{round(100 * 10212 / (3196*39), 2)}} & 6.6
    & \fpeval{round(100 * 11082 / (3196*39), 2)} & 5.0
    & \fpeval{round(100 * 11350 / (3196*39), 2)} & 0.1
    & \fpeval{round(100 * 15499 / (3196*39), 2)} & 1.4
    & \fpeval{round(100 * 10435 / (3196*39), 2)} & $\leq 0.1$
    & \fpeval{round(100 * 12486 / (3196*39), 2)} & 60\\
    
    & 19
    & \fpeval{round(100 * 4583 / (3196*39), 2)} & 2.0
    & \textbf{\fpeval{round(100 * 4083 / (3196*39), 2)}} & 8.8
    & \fpeval{round(100 * 8469 / (3196*39), 2)} & 10
    & \fpeval{round(100 * 6396 / (3196*39), 2)} & 0.1
    & \fpeval{round(100 * 13766 / (3196*39), 2)} & 2.7
    & \fpeval{round(100 * 4166 / (3196*39), 2)} & $\leq 0.1$
    & \fpeval{round(100 * 8432 / (3196*39), 2)} & 86\\

    \hline

    \multirow{3}{*}{Hepatis}
    
    & 15
    & \textbf{\fpeval{round(100 * 789 / (155*337), 2)}} & 0.1
    & \textbf{\fpeval{round(100 * 791 / (155*337), 2)}} & 0.3
    & \fpeval{round(100 * 947 / (155*337), 2)} & 50
    & \fpeval{round(100 * 842 / (155*337), 2)} & $\leq 0.1$
    & \fpeval{round(100 * 1113 / (155*337), 2)} & 0.1
    & \textbf{\fpeval{round(100 * 789 / (155*337), 2)}} & $\leq 0.1$
    & {\fpeval{round(100 * 1065 / (155*337), 2)}} & 20 \\
    
    & 39
    & \fpeval{round(100 * 554 / (155*337), 2)} & 0.2
    & \fpeval{round(100 * 555 / (155*337), 2)} & 0.8
    & \fpeval{round(100 * 765 / (155*337), 2)} & 150
    & \fpeval{round(100 * 639 / (155*337), 2)} & 0.2
    & \fpeval{round(100 * 1113 / (155*337), 2)} & 0.1
    & \textbf{\fpeval{round(100 * 551 / (155*337), 2)}} & $\leq 0.1$
    & {\fpeval{round(100 * 889 / (155*337), 2)}} & 34 \\
    
    & 77 
    & \textbf{\fpeval{round(100 * 347 / (155*337), 2)}} & 0.3
    & \textbf{\fpeval{round(100 * 345 / (155*337), 2)}} & 1.6
    & \fpeval{round(100 * 543 / (155*337), 2)}  & 240
    & \fpeval{round(100 * 446 / (155*337), 2)} & 0.2
    & \fpeval{round(100 * 1113 / (155*337), 2)} & 0.1
    & \fpeval{round(100 * 349 / (155*337), 2)} & $\leq 0.1$
    & {\fpeval{round(100 * 611 / (155*337), 2)}} & 60\\

    \hline

    \multirow{3}{*}{Lung}
    
    & 3
    & \fpeval{round(100 * 997 / (31*147), 2)} & $\leq 0.1$
    & \fpeval{round(100 * 981 / (31*147), 2)} & 1.7
    & \fpeval{round(100 * 1164 / (31*147), 2)} & 9.7
    & \fpeval{round(100 * 1149 / (31*147), 2)} & $\leq 0.1$
    & \fpeval{round(100 * 1260 / (31*147), 2)} & $\leq 0.1$
    & \fpeval{round(100 * 1195 / (31*147), 2)} & $\leq 0.1$
    & \textbf{\fpeval{round(100 * 966 / (31*147), 2)}} & 3.4 \\
    
    & 8 
    & \fpeval{round(100 * 773 / (31*147), 2)} & 0.1
    & \textbf{\fpeval{round(100 * 702 / (31*147), 2)}} & 12
    & \fpeval{round(100 * 1040 / (31*147), 2)} & 26
    & \fpeval{round(100 * 993 / (31*147), 2)} & $\leq 0.1$
    & \fpeval{round(100 * 1123 / (31*147), 2)} & $\leq 0.1$
    & \fpeval{round(100 * 950 / (31*147), 2)} & $\leq 0.1$
    & \fpeval{round(100 * 793  / (31*147), 2)} & 4.1 \\
    
    & 16
    & \fpeval{round(100 * 497 / (31*147), 2)} & 0.1
    & \textbf{\fpeval{round(100 * 492 / (31*147), 2)}} & 12
    & \fpeval{round(100 * 687 / (31*147), 2)} & 57
    & \fpeval{round(100 * 566 / (31*147), 2)} & 0.1
    & \fpeval{round(100 * 906 / (31*147), 2)} & 0.1
    & \fpeval{round(100 * 634 / (31*147), 2)} & $\leq 0.1$
    & \fpeval{round(100 * 607 / (31*147), 2)} & 4.6\\

    \hline

    \multirow{3}{*}{Soybean}
    
    & 10
    & \fpeval{round(100 * 3018 / (307*100), 2)} & 0.2
    & \textbf{\fpeval{round(100 * 2971 / (307*100), 2)}} & 3.6
    & \fpeval{round(100 * 3301 / (307*100), 2)} & 10
    & \fpeval{round(100 * 3317 / (307*100), 2)} & $\leq 0.1$
    & \fpeval{round(100 * 4082 / (307*100), 2)} & 0.1
    & \fpeval{round(100 * 3332 / (307*100), 2)} & $\leq 0.1$
    & \fpeval{round(100 * 3493 / (307*100), 2)}  & 17 \\
    
    & 25 
    & \fpeval{round(100 * 1791 / (307*100), 2)} & 0.3
    & \textbf{\fpeval{round(100 * 1660 / (307*100), 2)}} & 9.6
    & \fpeval{round(100 * 2560 / (307*100), 2)} & 26
    & \fpeval{round(100 * 2266 / (307*100), 2)} & 0.1
    & \fpeval{round(100 * 3389 / (307*100), 2)} & 0.3
    & \fpeval{round(100 * 1955 / (307*100), 2)} & 0.1
    & \fpeval{round(100 * 2739 / (307*100), 2)} & 27\\
    
    & 50 
    & \fpeval{round(100 * 655 / (307*100), 2)} & 0.4
    & \textbf{\fpeval{round(100 * 615 / (307*100), 2)}} & 12
    & \fpeval{round(100 * 2130 / (307*100), 2)} & 59
    & \fpeval{round(100 * 1308 / (307*100), 2)} & 0.2
    & \fpeval{round(100 * 2979 / (307*100), 2)} & 0.3
    & \fpeval{round(100 * 739 / (307*100), 2)} & 0.1
    & \fpeval{round(100 * 2010 / (307*100), 2)} & 42 \\

    \hline

    \multirow{3}{*}{Tumor}
    
    & 4
    & \fpeval{round(100 * 1385 / (339*44), 2)} & $\leq 0.1$
    & \textbf{\fpeval{round(100 * 1333 / (339*44), 2)}} & 0.1
    & \fpeval{round(100 * 1431 / (339*44), 2)} & 1.8
    & \fpeval{round(100 * 1460 / (339*44), 2)} & $\leq 0.1$
    & \fpeval{round(100 * 1623 / (339*44), 2)} & $\leq 0.1$
    & \fpeval{round(100 * 1429 / (339*44), 2)} & $\leq 0.1$
    & \fpeval{round(100 * 1406 / (339*44), 2)} & 6.0\\
    
    & 11 
    & {\fpeval{round(100 * 756 / (339*44), 2)}} & 0.1
    & \textbf{\fpeval{round(100 * 736 / (339*44), 2)}} & 0.2
    & \fpeval{round(100 * 1046 / (339*44), 2)} & 4.8
    & \fpeval{round(100 * 844 / (339*44), 2)}  & $\leq 0.1$
    & \fpeval{round(100 * 1337 / (339*44), 2)} & 0.1
    & \fpeval{round(100 * 838 / (339*44), 2)}  & $\leq 0.1$
    & \fpeval{round(100 * 1037 / (339*44), 2)}  & 8.7\\
    
    & 22 
    & \textbf{\fpeval{round(100 * 290 / (339*44), 2)}} & 0.1
    & \fpeval{round(100 * 311 / (339*44), 2)} & 0.3
    & \fpeval{round(100 * 857 / (339*44), 2)} & 9.9
    & \fpeval{round(100 * 504 / (339*44), 2)}  & 0.1
    & \fpeval{round(100 * 1217 / (339*44), 2)} & 0.2
    & \fpeval{round(100 * 336 / (339*44), 2)}  & $\leq 0.1$
    & \fpeval{round(100 * 661 / (339*44), 2)} & 11\\

    \hline

    \multirow{3}{*}{Votes}
    
    & 2
    & \fpeval{round(100 * 1244 / (435*17), 2)} & $\leq 0.1$
    & \textbf{\fpeval{round(100 * 1243 / (435*17), 2)}} & $\leq 0.1$
    & \fpeval{round(100 * 1335 / (435*17), 2)} & 0.4
    & \fpeval{round(100 * 2054 / (435*17), 2)} & $\leq 0.1$
    & \fpeval{round(100 * 1486 / (435*17), 2)} & $\leq 0.1$
    & \fpeval{round(100 * 2341 / (435*17), 2)} & $\leq 0.1$
    & \fpeval{round(100 * 1335 / (435*17), 2)} & 4\\
    
    & 4 
    & \fpeval{round(100 * 1015 / (435*17), 2)} & 0.1
    & \textbf{\fpeval{round(100 * 955 / (435*17), 2)}} & 0.2
    & \fpeval{round(100 * 1068 / (435*17), 2)} & 0.8
    & \fpeval{round(100 * 977 / (435*17), 2)}  & $\leq 0.1$
    & \fpeval{round(100 * 1298 / (435*17), 2)} & $\leq 0.1$
    & \fpeval{round(100 * 1463 / (435*17), 2)} & $\leq 0.1$
    & \fpeval{round(100 * 993 / (435*17), 2)}  & 4.8\\
    
    & 8 
    & \fpeval{round(100 * 640 / (435*17), 2)} & 0.2
    & \textbf{\fpeval{round(100 * 526 / (435*17), 2)}} & 0.3
    & \fpeval{round(100 * 949 / (435*17), 2)} & 1.5
    & \fpeval{round(100 * 707 / (435*17), 2)}  & $\leq 0.1$
    & \fpeval{round(100 * 1073 / (435*17), 2)} & 0.1
    & \fpeval{round(100 * 863 / (435*17), 2)}  & $\leq 0.1$
    & \fpeval{round(100 * 740 / (435*17), 2)}  & 5.1\\

    \hline

    \multirow{3}{*}{Zoo}
    
    & 3
    & \fpeval{round(100 * 355 / (101*28), 2)} & $\leq 0.1$
    & \textbf{\fpeval{round(100 * 274 / (101*28), 2)}} & $\leq 0.1$
    & \textbf{\fpeval{round(100 * 274 / (101*28), 2)}} & 0.9
    & \fpeval{round(100 * 337 / (101*28), 2)} & $\leq 0.1$
    & \fpeval{round(100 * 294 / (101*28), 2)} & $\leq 0.1$
    & \fpeval{round(100 * 345 / (101*28), 2)} & $\leq 0.1$
    & \textbf{\fpeval{round(100 * 274 / (101*28), 2)}} & 2.9\\
    
    & 7 
    & \fpeval{round(100 * 157 / (101*28), 2)} & $\leq 0.1$
    & {\fpeval{round(100 * 150 / (101*28), 2)}} & $\leq 0.1$
    & \fpeval{round(100 * 168 / (101*28), 2)} & 2.0
    & \fpeval{round(100 * 191 / (101*28), 2)} & $\leq 0.1$
    & \fpeval{round(100 * 214 / (101*28), 2)} & $\leq 0.1$
    & \fpeval{round(100 * 178 / (101*28), 2)} & $\leq 0.1$
    & \textbf{\fpeval{round(100 * 132 / (101*28), 2)}} & 3.7\\
    
    & 14
    & \textbf{\fpeval{round(100 * 63 / (101*28), 2)}} & $\leq 0.1$
    & \textbf{\fpeval{round(100 * 63 / (101*28), 2)}} & $\leq 0.1$
    & \fpeval{round(100 * 143 / (101*28), 2)} & 3.8
    & \fpeval{round(100 * 109 / (101*28), 2)} & $\leq 0.1$
    & \fpeval{round(100 * 161 / (101*28), 2)} & $\leq 0.1$
    & \textbf{\fpeval{round(100 * 63 / (101*28), 2)}} & $\leq 0.1$
    & \fpeval{round(100 * 88 / (101*28), 2)} & 3.2\\

    \hline

    \multirow{3}{*}{Balance}
    
    & 2
    & \textbf{\fpeval{round(100 * 2549 / (625*23), 2)}} & $\leq 0.1$
    & \textbf{\fpeval{round(100 * 2549 / (625*23), 2)}} & 0.1
    & \textbf{\fpeval{round(100 * 2549 / (625*23), 2)}} & 0.6
    & \textbf{\fpeval{round(100 * 2549 / (625*23), 2)}} & $\leq 0.1$
    & \fpeval{round(100 * 2733 / (625*23), 2)} & 0.1
    & \textbf{\fpeval{round(100 * 2549 / (625*23), 2)}} & $\leq 0.1$
    & \textbf{\fpeval{round(100 * 2549 / (625*23), 2)}} & 6 \\
    
    & 6 
    & \textbf{\fpeval{round(100 * 2049 / (625*23), 2)}} & $\leq 0.1$
    & \textbf{\fpeval{round(100 * 2049 / (625*23), 2)}} & 0.3
    & \fpeval{round(100 * 2157 / (625*23), 2)} & 1.4
    & \textbf{\fpeval{round(100 * 2049 / (625*23), 2)}} & $\leq 0.1$
    & \fpeval{round(100 * 2357 / (625*23), 2)} & 0.2
    & \textbf{\fpeval{round(100 * 2049 / (625*23), 2)}} & $\leq 0.1$
    & \textbf{\fpeval{round(100 * 2049 / (625*23), 2)}} & 8.0\\
    
    & 11 
    & \textbf{\fpeval{round(100 * 1424 / (625*23), 2)}} & $\leq 0.1$
    & \textbf{\fpeval{round(100 * 1424 / (625*23), 2)}} & 0.4
    & \fpeval{round(100 * 1728 / (625*23), 2)} & 2.5
    & \textbf{\fpeval{round(100 * 1424 / (625*23), 2)}} & $\leq 0.1$
    & \fpeval{round(100 * 2282 / (625*23), 2)} & 0.3
    & \textbf{\fpeval{round(100 * 1424 / (625*23), 2)}} & $\leq 0.1$
    & \textbf{\fpeval{round(100 * 1424 / (625*23), 2)}} & 10 \\

    \hline

    \multirow{3}{*}{Car}
    
    & 2 
    & {\fpeval{round(100 * 10310 / (1728*25), 2)}} & 0.1
    & {\fpeval{round(100 * 10310 / (1728*25), 2)}} & 1.6
    & \fpeval{round(100 * 10292 / (1728*25), 2)} & 0.5
    & {\fpeval{round(100 * 10310 / (1728*25), 2)}} & $\leq 0.1$
    & \fpeval{round(100 * 11401 / (1728*25), 2)} & 0.6
    & {\fpeval{round(100 * 10310 / (1728*25), 2)}} & $\leq 0.1$
    & \textbf{\fpeval{round(100 * 9984 / (1728*25), 2)}} & 12 \\
    
    & 6 
    & {\fpeval{round(100 * 8006 / (1728*25), 2)}} & 0.2
    & {\fpeval{round(100 * 8006 / (1728*25), 2)}} & 2.7
    & \fpeval{round(100 * 8298 / (1728*25), 2)} & 1.6
    & {\fpeval{round(100 * 8006 / (1728*25), 2)}} & $\leq 0.1$
    & \fpeval{round(100 * 10250 / (1728*25), 2)} & 1.6
    & \fpeval{round(100 * 8006 / (1728*25), 2)} & $\leq 0.1$
    & \textbf{\fpeval{round(100 * 7824 / (1728*25), 2)}} & 20\\
    
    & 12 
    & \textbf{\fpeval{round(100 * 4838 / (1728*25), 2)}} & 0.3
    & \textbf{\fpeval{round(100 * 4838 / (1728*25), 2)}} &  3.7
    & \fpeval{round(100 * 6203 / (1728*25), 2)} & 3.0
    & \textbf{\fpeval{round(100 * 4838 / (1728*25), 2)}} & $\leq 0.1$
    & \fpeval{round(100 * 8791 / (1728*25), 2)} & 2.5
    & \textbf{\fpeval{round(100 * 4838 / (1728*25), 2)}} & $\leq 0.1$
    & \fpeval{round(100 * 4912 / (1728*25), 2)} & 24\\
    
    
    \hline

    \multirow{3}{*}{Lympho}
    
    & 5
    & \fpeval{round(100 * 1184 / (148*54), 2)} & $\leq 0.1$
    & \textbf{\fpeval{round(100 * 1170 / (148*54), 2)}} & 0.1
    & \fpeval{round(100 * 1202 / (148*54), 2)} & 2.7
    & \fpeval{round(100 * 1272 / (148*54), 2)} & $\leq 0.1$
    & \fpeval{round(100 * 1385 / (148*54), 2)} & $\leq 0.1$
    & \fpeval{round(100 * 1217 / (148*54), 2)} & $\leq 0.1$
    & \fpeval{round(100 * 1198 / (148*54), 2)} & 5\\
    
    & 13
    & \textbf{\fpeval{round(100 * 788 / (148*54), 2)}} & $\leq 0.1$
    & \fpeval{round(100 * 790 / (148*54), 2)} & 0.3
    & \fpeval{round(100 * 998 / (148*54), 2)} & 7.0
    & \fpeval{round(100 * 899 / (148*54), 2)} & $\leq 0.1$
    & \fpeval{round(100 * 1234 / (148*54), 2)} & $\leq 0.1$
    & \fpeval{round(100 * 842 / (148*54), 2)} & $\leq 0.1$
    & \fpeval{round(100 * 969 / (148*54), 2)} & 9\\
    
    & 27
    & \textbf{\fpeval{round(100 * 286 / (148*54), 2)}} & 0.1
    & \fpeval{round(100 * 305 / (148*54), 2)} & 0.4
    & \fpeval{round(100 * 855 / (148*54), 2)} & 15
    & \fpeval{round(100 * 471 / (148*54), 2)} & $\leq 0.1$
    & \fpeval{round(100 * 1035 / (148*54), 2)} & 0.1
    & \fpeval{round(100 * 311 / (148*54), 2)} & $\leq 0.1$
    & \fpeval{round(100 * 688 / (148*54), 2)} & 11\\
    
    
    \hline

    \multirow{3}{*}{Nursery}
    
    & 3
    & \fpeval{round(100 * 95040 / (12960*31), 2)} & 2.9
    & \fpeval{round(100 * 90108 / (12960*31), 2)} & 170
    & \textbf{\fpeval{round(100 * 89844 / (12960*31), 2)}} & 1.8
    & \fpeval{round(100 * 95040 / (12960*31), 2)} & 0.1
    & \fpeval{round(100 * 99696 / (12960*31), 2)} & 16
    & \fpeval{round(100 * 90720 / (12960*31), 2)} & $\leq 0.1$
    & \fpeval{round(100 * 90108 / (12960*31), 2)} & 110 \\
    
    & 8 
    & \fpeval{round(100 * 73440 / (12960*31), 2)} & 6.4
    & \textbf{\fpeval{round(100 * 68004 / (12960*31), 2)}} & 560
    & \fpeval{round(100 * 71154 / (12960*31), 2)} & 4.5
    & \fpeval{round(100 * 73440 / (12960*31), 2)} & 0.2
    & \fpeval{round(100 * 93284 / (12960*31), 2)} & 25
    & \fpeval{round(100 * 69120 / (12960*31), 2)} & 0.1
    & \fpeval{round(100 * 69084 / (12960*31), 2)} & 160\\
    
    & 15 
    & \textbf{\fpeval{round(100 * 39210 / (12960*31), 2)}} & 13
    & \fpeval{round(100 * 39456 / (12960*31), 2)} & 640
    & \fpeval{round(100 * 48408 / (12960*31), 2)} & 8.5
    & \fpeval{round(100 * 39924 / (12960*31), 2)} & 0.5
    & \fpeval{round(100 * 80805 / (12960*31), 2)} & 33
    & \fpeval{round(100 * 39924 / (12960*31), 2)} & 0.1
    & \fpeval{round(100 * 43438 / (12960*31), 2)} & 230 \\

    \hline

    \multirow{3}{*}{Phishing}
    
    & 3
    & \fpeval{round(100 * 7794 / (1353*26), 2)} & 0.3
    & \textbf{\fpeval{round(100 * 7414 / (1353*26), 2)}} & 0.9
    & \fpeval{round(100 * 7598 / (1353*26), 2)} & 0.9
    & \fpeval{round(100 * 8227 / (1353*26), 2)} & $\leq 0.1$
    & \fpeval{round(100 * 8281 / (1353*26), 2)} & 0.2
    & \fpeval{round(100 * 8620 / (1353*26), 2)} & $\leq 0.1$
    & \fpeval{round(100 * 7527 / (1353*26), 2)} & 12\\
    
    & 6 
    & \fpeval{round(100 * 6276 / (1353*26), 2)} & 0.5
    & \textbf{\fpeval{round(100 * 6045 / (1353*26), 2)}} & 1.4
    & \fpeval{round(100 * 6569 / (1353*26), 2)} & 1.7
    & \fpeval{round(100 * 6249 / (1353*26), 2)} & $\leq 0.1$
    & \fpeval{round(100 * 7666 / (1353*26), 2)} & 0.4
    & \fpeval{round(100 * 6757 / (1353*26), 2)} & $\leq 0.1$
    & \fpeval{round(100 * 6216 / (1353*26), 2)} & 15\\
    
    & 13 
    & \fpeval{round(100 * 3135 / (1353*26), 2)} & 0.7
    & \textbf{\fpeval{round(100 * 3066 / (1353*26), 2)}} & 1.9
    & \fpeval{round(100 * 5085 / (1353*26), 2)} & 3.6
    & \fpeval{round(100 * 3681 / (1353*26), 2)} & $\leq 0.1$
    & \fpeval{round(100 * 6794 / (1353*26), 2)} & 0.6
    & \fpeval{round(100 * 3411 / (1353*26), 2)} & $\leq 0.1$
    & \fpeval{round(100 * 4114 / (1353*26), 2)} & 19\\

    \hline

    \multirow{3}{*}{Student}
    & 18
    & \textbf{\fpeval{round(100 * 5864 / (395*176), 2)}} & 0.4
    & \fpeval{round(100 * 5898 / (395*176), 2)} & 31
    & \fpeval{round(100 * 6591 / (395*176), 2)} & 31
    & \fpeval{round(100 * 5891 / (395*176), 2)} & 0.1
    & \fpeval{round(100 * 8180 / (395*176), 2)} & 0.2
    & \fpeval{round(100 * 5910 / (395*176), 2)} & 0.2
    & \fpeval{round(100 * 6767 / (395*176), 2)} & 58 \\
    
    & 44 
    & \textbf{\fpeval{round(100 * 3387 / (395*176), 2)}} & 0.6
    & \fpeval{round(100 * 3397 / (395*176), 2)} & 69
    & \fpeval{round(100 * 5232 / (395*176), 2)} & 93
    & \fpeval{round(100 * 3521 / (395*176), 2)} & 0.2
    & \fpeval{round(100 * 7317 / (395*176), 2)} & 0.3
    & \fpeval{round(100 * 3400 / (395*176), 2)} & 0.2
    & \fpeval{round(100 * 5540 / (395*176), 2)} & 100\\
    
    & 88
    & \textbf{\fpeval{round(100 * 1198 / (395*176), 2)}} & 0.8
    & \fpeval{round(100 * 1234 / (395*176), 2)} & 71
    & \fpeval{round(100 * 4195 / (395*176), 2)} & 200
    & \fpeval{round(100 * 1610 / (395*176), 2)} & 0.2
    & \fpeval{round(100 * 6340 / (395*176), 2)} & 0.6
    & \fpeval{round(100 * 1227 / (395*176), 2)} & 0.3
    & \fpeval{round(100 * 3804 / (395*176), 2)} & 180\\
    
    \hline

    \multirow{3}{*}{Tictactoe}
    
    & 3
    & \fpeval{round(100 * 7660 / (958*28), 2)} & $\leq 0.1$
    & \fpeval{round(100 * 7660 / (958*28), 2)} & 1.6
    & \fpeval{round(100 * 7607 / (958*28), 2)} & 0.8
    & \fpeval{round(100 * 7660 / (958*28), 2)} & $\leq 0.1$
    & \fpeval{round(100 * 7916 / (958*28), 2)} & 0.2
    & \fpeval{round(100 * 7660 / (958*28), 2)} & $\leq 0.1$
    & \textbf{\fpeval{round(100 * 7444 / (958*28), 2)}} & 14\\
    
    & 7 
    & \fpeval{round(100 * 6068 / (958*28), 2)} & 0.2
    & \fpeval{round(100 * 6062 / (958*28), 2)} & 3.1
    & \fpeval{round(100 * 6175 / (958*28), 2)} & 2.0
    & \fpeval{round(100 * 6068 / (958*28), 2)} & $\leq 0.1$
    & \fpeval{round(100 * 6846 / (958*28), 2)} & 0.3
    & \fpeval{round(100 * 6068 / (958*28), 2)} & $\leq 0.1$
    & \textbf{\fpeval{round(100 * 5951 / (958*28), 2)}} & 14\\
    
    & 14 
    & \textbf{\fpeval{round(100 * 3588 / (958*28), 2)}} & 0.2
    & \textbf{\fpeval{round(100 * 3588 / (958*28), 2)}} & 3.6
    & \fpeval{round(100 * 4436 / (958*28), 2)} & 4.0
    & \textbf{\fpeval{round(100 * 3588 / (958*28), 2)}} & 0.1
    & \fpeval{round(100 * 6085 / (958*28), 2)} & 0.4
    & \textbf{\fpeval{round(100 * 3588 / (958*28), 2)}} & $\leq 0.1$
    & \fpeval{round(100 * 3901 / (958*28), 2)} & 20\\

    \hline

    \multirow{3}{*}{Wine}
    
    & 18 
    & \textbf{\fpeval{round(100 * 2117 / (178*1279), 2)}} & 0.3
    & \textbf{\fpeval{round(100 * 2117 / (178*1279), 2)}} & 1.4
    & \textbf{\fpeval{round(100 * 2117 / (178*1279), 2)}}  & 230
    & \textbf{\fpeval{round(100 * 2117 / (178*1279), 2)}}  & 0.1
    & \fpeval{round(100 * 2482 / (178*1279), 2)} & 0.1
    & \textbf{\fpeval{round(100 * 2117 / (178*1279), 2)}} & 0.4
    & \fpeval{round(100 * 2240 / (178*1279), 2)} & 90\\
    
    & 44 
    & \textbf{\fpeval{round(100 * 1779 / (178*1279), 2)}} & 0.5
    & \textbf{\fpeval{round(100 * 1779 / (178*1279), 2)}} & 3.0
    & \textbf{\fpeval{round(100 * 1781 / (178*1279), 2)}} & 530
    & \textbf{\fpeval{round(100 * 1779 / (178*1279), 2)}} & 0.1
    & \fpeval{round(100 * 2482 / (178*1279), 2)} & 0.1
    & \textbf{\fpeval{round(100 * 1779 / (178*1279), 2)}} & 0.8
    & \fpeval{round(100 * 1876 / (178*1279), 2)} & 160\\
    
    & 89 
    & \textbf{\fpeval{round(100 * 1194 / (178*1279), 2)}} & 0.7
    & \textbf{\fpeval{round(100 * 1194 / (178*1279), 2)}} & 6.0
    & \fpeval{round(100 * 1196 / (178*1279), 2)} & 1000
    & \textbf{\fpeval{round(100 * 1194 / (178*1279), 2)}} & 0.4
    & \fpeval{round(100 * 2482 / (178*1279), 2)} & 0.1
    & \textbf{\fpeval{round(100 * 1194 / (178*1279), 2)}} & 1.5
    & \fpeval{round(100 * 1246 / (178*1279), 2)} & 180\\
    
    
    \hline

    Mush-
    & 11 
    & \fpeval{round(100 * 65681 / (8124*111), 2)} & 50
    & \textbf{\fpeval{round(100 * 60484 / (8124*111), 2)}} & 1200
    & \fpeval{round(100 * 78754 / (8124*111), 2)} & 60
    & \fpeval{round(100 * 68362 / (8124*111), 2)} & 0.7
    & \fpeval{round(100 * 86303 / (8124*111), 2)} & 23
    & \fpeval{round(100 * 67890 / (8124*111), 2)} & 0.3
    & \fpeval{round(100 * 74026 / (8124*111), 2)} & 510\\room
    
    & 28  
    & \fpeval{round(100 * 35312 / (8124*111), 2)} & 55
    & \textbf{\fpeval{round(100 * 31290 / (8124*111), 2)}} & 2500
    & \fpeval{round(100 * 68000 / (8124*111), 2)}  & 150
    & \fpeval{round(100 * 40182 / (8124*111), 2)}  & 0.7
    & \fpeval{round(100 * 75411 / (8124*111), 2)}  & 25
    & \fpeval{round(100 * 33844 / (8124*111), 2)} & 0.4
    & \fpeval{round(100 * 59506 / (8124*111), 2)} & 600\\

    & 55  
    & \fpeval{round(100 * 14244 / (8124*111), 2)} & 58
    & \textbf{\fpeval{round(100 * 8926 / (8124*111), 2)}} & 2800
    & \fpeval{round(100 * 62047 / (8124*111), 2)}  & 280
    & \fpeval{round(100 * 23074 / (8124*111), 2)}  & 1.6
    & \fpeval{round(100 * 65968 / (8124*111), 2)}  & 41
    & \fpeval{round(100 * 9868 / (8124*111), 2)} & 1.2
    & \fpeval{round(100 * 41689 / (8124*111), 2)} & 1100\\
    

    
    %

    \hline
    \hline

    \end{tabular}
    \end{table}

\end{document}